\documentclass[article]{elsarticle}                           

\usepackage{lineno,hyperref}
\modulolinenumbers[5]
\usepackage{times}

\usepackage{soul}
\usepackage{url}
\usepackage[utf8]{inputenc}
\usepackage[small]{caption}
\usepackage{graphicx}
\usepackage{amsmath}
\usepackage{booktabs}
\begin{document}

\begin{frontmatter}
\title{’Controlling by Showing: i-Mimic’: A Video-based Method to Control Robotic Arms}

\author[mymainaddress]{Debarati B. Chakraborty\corref{mycorrespondingauthor}}
\cortext[mycorrespondingauthor]{Debarati B. Chakraborty}
\ead{debarati.earth@gmail.com}

\author[mysecondaryaddress]{Mukesh Sharma}
\ead{sharma.15@iitj.ac.in}

\author[mysecondaryaddress]{Bhaskar Vijay}
\ead{vijay.2@iitj.ac.in}

\address[mymainaddress]{Dept. of Computer Science and Engineering, Indian Institute of Technology, Jodhpur, India}
\address[mysecondaryaddress]{Department of Mechanical Engineering, Indian Institute of Technology, Jodhpur, India}






\begin{abstract}

A novel concept of vision-based intelligent control of robotic arms is developed here in this work. This work enables the controlling of robotic arms motion only with visual inputs, that is, controlling by showing the videos of correct movements. This work can broadly be sub-divided into two segments. The first part of this work is to develop an unsupervised vision-based method to control robotic arm in 2-D plane, and the second one is with deep CNN in the same task in 3-D plane. The first method is unsupervised, where our aim is to perform mimicking of human arm motion in real-time by a manipulator. Mimicking, here involves a series of steps, namely, tracking the motion of the arm in videos, estimating motion parameters, and replicating the motion parameters in the robot. We developed a network, namely the vision-to-motion optical network (DON), where the input should be a video stream containing hand movements of human, the the output would be out the velocity and torque information of the hand movements shown in the videos. The output information of the DON is then fed to the robotic
arm by enabling it to generate motion according to the real hand videos. The method has been tested with both live-stream video feed as well as on recorded video obtained from a monocular camera. Besides, the DON also enables a method to behave intelligently by predicting the trajectory of human hand even if the hand gets occluded for some time with varying degree of occlusion. This is why the mimicry of the arm incorporates some intelligence to it and becomes intelligent mimic (i- mimic).
Alongside the unsupervised method another method has also been developed deploying the deep neural network technique with CNN (Convolutional Neural Network) to perform the mimicking, where labelled datasets are used for training. The same dataset, as used in the unsupervised DON-based method, is used in the deep CNN method, after manual annotations. Both the proposed methods are validated with off-line as well as with on-line video datasets in real-time, enabling the robotic arm in ’i- mimic’ and thereby showing the effectiveness of the proposed method. The entire methodology is validated with real-time 1-link and simulated n-link manipulators alongwith suitable comparisons.
\end{abstract}

\end{frontmatter}
\section{Introduction}
Robotic arms are used to perform mechanical tasks in industries over
decades. It was mainly used for performing repetitive tasks in the
industries to cut down the labor cost \cite{Basu_17,Ali_18}.
Normally, robotic arms are quite complex with five or more degree of
freedom as it aims to perform human tasks. Recently, the
application of robotic arms in conducting the domestic work has
drawn attention. Controlling of these robotic arms to perform different tasks is still a major issue to be addressed. Here in this work we have defined a new concept of controlling robotic arms only with visual information, that is, the motion of different parts of the robotic arms could be controlled by providing according human hand movement videos as the input to those arms. 

The entire method could be subdivided into two parts.
In the first part of the this work we aim to find out a simple solution for unsupervised
controlling the robotic arm only with visual information. Here we aim to deal with the issues of
i) unavailability of sufficient training datasets, ii) domain adaptation and
iii) economic cost. On the way to search for a solution of the two
initial issues we concluded that the teaching/ training part should be
removed. But how could it be controlled then? 'Mimic' is the
solution that stroked in our mind. The controlling could only be achieved by showing the arm the desired movement and making it enabled
to follow it. Visual mimic-based controlling of 1-link and n-linked robotic arms is the
primary contribution of this work. The mechanism of this set-up is
quite simple and the 1-link manipulator is developed only by the authors. Either the recorded or real-time video could be shown to the arm to
achieve the control. It should be noted that the real-time testing of this
'mimic' with robotic arm is in a very primary level where the arm is a simple 1-link
robotic manipulator with a degree of freedom of 120 degrees. The rest of the tests are conducted in the form of simulation, where
the simulated arm is able to mimic the motion of a single joint (solder or
elbow) or real hand.

Another contribution in this part of the work is development of the vision-to-motion optical network (DON) to
process the optical flow information of the input video and
to convert it into the physical force, to be fed the robotic arm.
The proposed DON is different from the existing deep networks in the following manners: i) it does not require any
labeled data set or manual intervention, ii) it does neither
require background estimation or a large number of input
frames for training, iii) the functioning of the intermediate
layers are simple which enables computational gain iv) all
the intermediate layers are not active simultaneously; some
layer gets activated depending on the values of the outputs
of its previous layer v) the single network can perform both
estimation and prediction and vi) it produces torque and angular velocity as the output.

In the second part of the work we focus on implementation rather than definition. We implemented deep neural network with CNN and Refiner Network for the purpose of i-mimic. To deal with unavailability of large set of  labelled data, we used a CNN network whose output are further fed into the refinement network that smoothens the final output and enables to interpolate to a larger range. The deep CNN is proven to be less effective with 2-D n-link manipulators, but, it performs better in 3-D plane with n-link manipulators.  

The rest of the article is organised as follows. Sec. \ref{Background Research} presents the background research,  Sec. \ref{DON} describes the layer-wise formulation of vision-to-motion optical network (DON). The architecture of CNN and Refinement Network, customized loss function, dataset, training are described in Sec. \ref{DNN}. The experimental set-ups with four different variations in experimental studies are described in Sec. \ref{exps}. The real-time experimental results tested under different scenarios like without occlusion, with low occlusion, with high occlusion, and with multiple joints are given in Sec. \ref{Res} along with parameter section and comparative study. The overall conclusion of this work with its future scope are discussed in Sec. \ref{con}.

\section{Background Research}\label{Background Research}

Vision-based robotics to serve domestic purpose has drawn the
attention of researchers in several areas. Most of the
approaches developed so far for this purpose implied
training the system through labeled data, i.e., with supervised or
semi-supervised learning. Automated driving \cite{Geiger_13},
grasping \cite{Levine_18} and block stacking \cite{Hundt_18} are among
few the applications where this kind of learning were
used. But gathering adequate amount of labeled data for training is a
challenging issue for this type of approaches.

Semi-supervised learning or reinforcement learning has appeared to be
the substitute of supervised learning in recent literature's where
the training is carried out with less amount of labeled data or
weakly labeled data. Rusu \emph{et al.} \cite{Rusu_17} used learning
with progressive network for Jaco robot gripping to have a faster
algorithm with less amount of training data. KUKA IIWA robot
grasping with deep network and domain adaptation was developed by
Bousmalis \emph{et al.} \cite{Bousmalis_18}. A method of training
with weakly labeled images with adaptation from real world to
simulation using a PR2 robot was proposed by Tzeng \emph{et al.}
\cite{Tzeng_17}. Zuo \emph{et al.} \cite{Zuo_19} came up with a
solution of semi supervised method of 3D pose estimation where the
training was carried out in a virtual environment. Its real world
implementation was done after domain adaptation. Domain adaptation
is another challenge while semi-supervised/ reinforcement learning
is carried out. There are many rich works carried out so far to
deal with this problem. Domain adaptation with back-propagation by
inducing an 'inverse-gradient layer' to the deep network was
formulated by Ganin and Lempitsky \cite{Ganin_15}. In another work
Ganin \emph{et al.} \cite{Ajakan_16} came up with a solution of
carrying out the training and testing of the network with the
features that are non-discriminative and domain invariant for
training and test data. Bousmalis \emph{et al.} \cite{Bousmalis_16}
came up with another solution of identifying the unique feature of
each domain to extract out the common features in the domains. They
have recently developed another way of domain adaptation with
simultaneous simulation \cite{Bousmalis_18}. Sing \emph{et al.}
\cite{Sing_17} demonstrated that passively collected data can be
paired with interaction data to learn visual representations for
end-to-end control policies that generalize substantially better to
unseen environments. However, less amount of labeled data or
synthetically labeled data are always required in all of the
aforementioned approaches.

Economic cost of a robotic arm controller is another major issue to
be dealt with to make the robotic arms be implementable for domestic
purpose. The controlling of the robotic arms are normally carried
out with multiple sensors which make it more costly. The robotic
arms like PR2 \cite{Tzeng_17}, Jaco \cite{Rusu_17} or KUKA IIWA
\cite{Bousmalis_18} costs around USD 20,000/- to USD 50,000/-. A
pocket friendly robotic arm is developed recently \cite{Zuo_19} but
but it still have the issues of synthetic labeled data and domain
adaptation.

\section{DON: Vision-to-Motion Optical Network}\label{DON}

Here we developed a network based on the information of optical flow
from frame-to- frame of a the input hand movement video sequence.
The major challenges that are to be addressed in the task of
'i-mimic' are: unavailability of sufficient number of labelled data,
computation time and the lag between input-video to robotic-arm
gestures. The proposed deep flow network is able to minimize all
these parameters simultaneously. First of all, no labelled data is
required here and the process is fully automatic. The computation
complexity of this method is quite low and there the lag is as less
as around ten frames here.

\begin{figure*}
    \centering
    \includegraphics[width=5.5in]{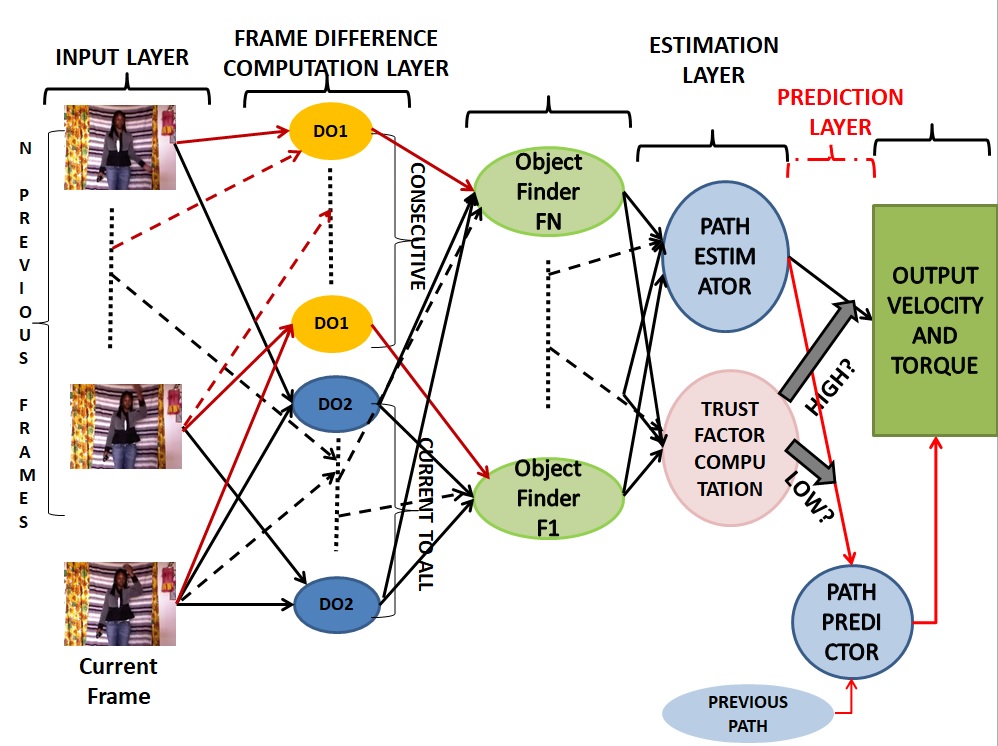}
    \caption{Architecture of Deep Flow Network}
    \label{DF-Arch}
\end{figure*}

The vision-to-motion optical network is a layer-wise network with multiple
layers between the input and output layer. Different feature of the
optical flow information is process in different layers of this
network. The layer-wise architecture is shown in Fig. \ref{DF-Arch}.  It can be noticed from the diagram that the videos
are fed in the input layer of the network, whereas we get velocity and torque values that generates the physical motion
at the output layer.
That is why it is named as a 'vision-to-motion optical network network'. The output of
the previous layer is the input to the next layer. All the layers of
the network may not be active at a time. Rather, the activation of
some layer of the network is dependent on the outputs of its
previous layer. The layer-wise working principles of this network
are described in details in the following sections.

\subsection{Layer 1: Input Layer}
The video sequence is the input that is fed to the network. But it
is not the entire sequence that is given as the input at a time since
a on-line processing is going on here with the input video frames.
As we have already stated that the proposed method is unsupervised,
therefore the output is to be produced only by automated processing
of input data. Here, the frame that gets generated on the current
instant, say, at the instant $t$ is fed to the network along-with
$N$-number of previous frames that gets generated earlier to the
current frame. Let the current frame be denoted as $f_t$here. The
frames generated in the earlier instances be denoted by $f_{t-1},
f_{t-2},...,f_{t-N}$. Therefore, the input layer contains the
frames: $f_t, f_{t-1},...,f_{t-N}$.

\subsection{Layer 2: Frame Difference Computation Layer}
The network is supposed to deal with the optical flow information.
In case of the video sequences that we are dealing with is captured
by static cameras. Therefore the changed information from frame to
frame reflects the optical flow of the sequence. Here two types of
differences are computed here in this layer. That is why two
different colored of nodes (DO1 and DO2) are shown there in Fig.
\ref{DF-Arch}. There are total $N+N=2N$ number of nodes present
there in layer 2. The difference operation carried out in DO1 the
difference between consecutive frames ($\delta1$) given in Eqn
(\ref{f1}). The difference between the current to all its previous
frames ($\delta2$) is carried out in Eqn (\ref{f2}).

\begin{equation}\label{f1}
\delta1_p=|f_{t-p}-f_{t-(p-1)}|:p=0,...,N-1
\end{equation}
\begin{equation}\label{f2}
\delta2_p=|f_{t}-f_{t-p}|:p=1,...,N
\end{equation}

Therefore $N$ number of binarized $\delta1$ and $\delta2$ frames,
i.e., $2N$ number of difference frames in total are the output of
this layer. $\delta1_p$ are the binarized outputs from the nodes of
type DO1, whereas, $\delta2_p$ are the output from DO2 type of
nodes.

\subsection{Layer 3: Object Identification Layer}\label{OIL}
The third layer of this network is developed to find out the
locations and the shape of the moving hand in all the $N$-number of
previous frames. As it can be observed from Fig. \ref{DF-Arch} that
this layer contains $N$ number of nodes, labeled as $Object Finder
F1,..., Object Finder FN$. The input fed to a certain node $Object
Finder Fp$ are: ${\delta2_p}:p=1,...,N$ and $\delta2_p$. That is,
all the DO2- difference frames and only $p^{th}$ DO1 difference
frame are the input to the said node of layer 3. Let the location of
the moving object segment in the $p^{th}$-frame be represented by
$l_p$. The operation that is carried out in each node of the third
layer is given by the Eqn. (\ref{OL}).

\begin{equation}\label{OL}
l_p=(\cup\ _{p=1}^{N}\delta2_p)\cap \delta1_p
\end{equation}
Please note that the union of $\delta2_p\forall p=1,...,N$ is taken
here to have the entire moving obeject region as a subset of that
union and intersection of it to that of $\delta1_p$ is carried out
to extract out the obvious moving region in the $p^{th}$-frame. The
pixels those belong to the set $l_p$ are in the region that
definitely belong to the moving object in the $p^{th}$ frame. For
the sake of simplicity, here in this work we consider only the
skeleton and the locations of the corner pixels of the $l_p$ (moving
hand) to be the output from each node of this layer as we need to
find out the angular velocity and torque from the hand movement
video.

\subsection{Layer 4: Estimation Layer}
This layer contains two nodes and the operations and functioning of
these two nodes are different from each other. The location of the
object in $N$-number of frames are the input to this layer. Two
types of estimations are performed simultaneously in this layer with
the two nodes. The path estimation node gives the probable
trajectory of the moving object as the output whereas the trust
factor estimator node computes the reliability of the estimated
path. The output of trust factor estimator node determines the
activation of the next layer, i.e., prediction layer. The working
principles of the two nodes in the forth layer are described below.

\subsubsection{Path Estimator}
As discussed before the prediction of probable trajectory of the
object is carried out here. This is done by computing the optical
velocity and acceleration of the moving object from frame-to-frame
displacement. Let $\varsigma_p$ be the location $l_p$ (see Eqn.
(\ref{OL})) in the $p^{th}$-frame. Then the velocity ($v_p$) and
acceleration ($a_p$) of that object are computed according to Eqn.
(\ref{velAcc}). The velocity and acceleration values for all the $N$
frames are stored in the sets $V$ and $A$ respectively.

\begin{align}\label{velAcc}
  {V}&=\{v_p: v_p=\varsigma_p-\varsigma_{p-1}\forall p=1,...,N\}\\
A&=\{a_p: a_p=v_p-v_{p-1}\forall p=2,...,N\}
\end{align}

Please note that signed difference between the locations and
velocity are taken while computing $v_p$ and $a_{p}$ in Eqn.
(\ref{velAcc}). It is known from Sec. \ref{OIL} that the input
$\varsigma_p$ could be a scalar or vector component based on the
type of object representation. But the two components $v_p$ and
$a_p$ should always be a vector since these components contain both
magnitude and signs. Consideration of those signs helps in the
incorporation of the information of the direction and the change in
the direction of the moving hand.

Here the robotic arm with revolute joint is supposed to mimic the movement of the arm shown in real time or recorded video. Therefore, the movement of the arm is always
supposed to be circular in nature with respect to any joint (e.g.
elbow) with maximum 180 degree of freedom. This phenomenon is kept
in mind and the determination of the radius w.r.t the angular motion
is computed by measuring the length of the skeleton of the arm. Let
the skeleton of the moving part of the arm be of length $r$. The
angular velocity ($\omega_p$) and torque ($\theta_p$) are then
computed as: $$\omega_p=\frac{v_p}{r}$$ and
$$\theta_p = I\alpha_p$$
where $I$ is mass moment of inertia of the manipulator arm and  $\alpha_p$ is the angular acceleration computed as:
$$\alpha_p = \frac{a_p}{r}$$
For any given one-link manipulator the algorithm computes $\alpha_p$ and having $I$ of the manipulator one can compute the torque required to be applied to at the joint.

\subsubsection{Trust Factor Computation}
The working principle of this particular node is different from any
other nodes present there in a network. It takes input from the
previous layer but does not transmit its output to the next layer.
Instead the output from this layer determines which layer should be
the fifth layer of this network. That is the which path should be
followed by the output information from the path estimator node is
decided with the output of this node. Since, only motion of the
moving hand of a static person is considered here, it can be assumed
that the size of the moving object will remain almost the same
throughout the sequence. This assumption is applied during
formulation of the trust factor. Let there be $M_p$ be the region of
$l_p$ (see Eqn. (\ref{OL})) in the $p^{th}$-frame. Let, the set
$\{S\}$ the regions of the object in all the $N$ frames and the set
$\{S_d\}$ contains the values of change in regions. Those are
computed according to Eqn. (\ref{RegS}).

\begin{align}\label{RegS}
  {S}&=\{M_p: p=1,...,N\}\nonumber\\
S_d&=\{c_p: c_p=|M_1-M_p| \forall p=2,...,N\}
\end{align}

The trust factor ($\eta$) is computed as:
\begin{equation}\label{TF}
    \eta=1-\frac{max(S_d)}{max(S)}
\end{equation}

In Eqn. (\ref{TF}) max(.) represents the element with maximum
magnitude present in a set. Physically, the effectiveness of
measuring the $\eta$ is in determining the amount of occlusion took
place over the moving object. If huge amount of occlusion is present
there for some frames, then the estimation with those frame may lead
to a wrong trajectory. Therefore, prediction should be carried out
from the previous set of information and ignoring the wrong
(occluded) visual information. That is why the activation of the
prediction layer is necessary in this scenario. The path leading to
prediction layer gets only activated if the value of $\eta$ is low.

\subsection{Layer 5: Prediction Layer}
There is only one node in this layer. But, the input fed to this
layer is not only from the previous layer, but the output of layer 4
of the previous execution of the network is also an input here. Please
note that this layer can not be active in the first execution of the
network, but from the second execution onward it could get activated any
time. Here the velocity and acceleration values from the frames
without occlusion, or with minimal occlusion are considered. There
inputs that are provided to this node are: i) the velocity and
acceleration information (sets $V$ and $A$) from the previous execution,
ii) the velocity and acceleration information ($V$ and $A$ from Eqn.
(\ref{velAcc})) from the previous layer and iii) Object regions and
change in the regions ($S$ and $S_d$ from Eqn. (\ref{RegS})). Let
the velocity and acceleration from the previous be denoted here as
$\Tilde{V}$  and $\Tilde{A}$. We only consider the information of
the frames for with $c_p<0.05Xmax(S)$ ($C_p$ is as defined in Eqn.
(\ref{RegS})). That is, frames maximum with $5\%$ change in object
size will be taken into account. Let $k$ number of frames out of the
$N$ frames failed to satisfy the criterion. Then only $N-k$ elements
from the sets $V$ and $A$ will be taken by merging it with the sets
$\Tilde{V}$  and $\Tilde{A}$ respectively. Therefore, the new sets
will be $V^k=\{\Tilde{V}|V(1:N-k)\}$ and
$A^k=\{\Tilde{A}|A(1:N-k)\}$ with $N+N-k=2N-k$ number of elements in
each set. Now we need to predict the information from the
$(N-k+1)^{th}$ frame onward. As it is known, the consecutive
difference between the elements of $V^k$ forms $A^k$, i.e., $A^k$
could be said the first order derivative of $V^k$. We can similarly
compute the second order derivative of $V^k$ or the first order
derivative of $A^k$ and represent it by $A^{k'}$. Now, the
$(2N-k+1)^{th}$ element of the sets $V^k$ and $A^k$ will be
approximated as:
\begin{align}\label{VAPr}
    v_{2N-k+1}=v_{2N-k}+a_{2N-k-1}\nonumber\\
    a_{2N-k+1}=a_{2N-k}+a'_{2N-k-1}
\end{align}
In Eqn. \ref{VAPr} the symbols $V_p$, $a_p$ and $a'_p$ represents
the $p^{th}$ element of the sets $V^k$, $A^k$ and $A^{k'}$
respectively. The element will get inserted to the sets $V^k$ and
$A^k$ as the $(2N-k+1)^{th}$ elements of them. The process will be
repeated and the next element will be approximated. The process will
continue until the set is going to have $2N$ number of elements.
Once it is done, the last $N$ elements of $V^k$ and $A^k$ will be
stored in the sets $V$ and $A$ respectively and will be given as the
output to the output layer.

The experimentation those are carried out with this proposed DON are described in the following section. Please note that, one additional layer, namely object regression layer to this network is introduced while working with multiple joints. It is shown in Fig. \ref{TR}.

\section{DNN:Deep Neural Network}\label{DNN}
In our second approach, we designed Convolutional Neural Network(CNN) for which the input corresponds to the image(frames of video stream) and labels correspond to coordinates$(x, y)$ of  joints(shoulder, elbow and wrist joints). We have used two networks for our purpose to get the joints coordinates from which further joint angles, joint velocities and and joint torque can be obtained in terms of pixel coordinates which are mapped to real value using the mapping function or mapping factor. In \cite{toshev2014deeppose} after using DNN based regression a DNN based refiner network was added, which takes cropped image around prediction as input to improve the prediction but we used a different approach by mapping a simple neural network to refine the predictions of the DNN based regression network. Our approach reduces the training time of the network with similar accuracy for our dataset.

\subsection{Convolutional Neural Network(CNN)}\label{cnn-network}
\begin{figure}
    \centering
    \includegraphics[width=5 in]{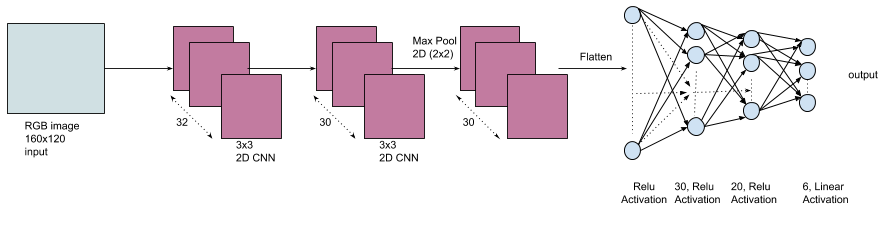}
    \caption{CNN Network}
    \label{CNN-Netork}
\end{figure}

The architecture of CNN is shown in Fig. \ref{CNN-Netork}. The input to this network is the images(from a stream of video feed) and labels as the pixel coordinates$(x, y)$ of the shoulder joint, elbow joint and wrist joint. The network has six hidden layers wherein there are 2 Convolutional layers followed by max pooling and flatten. Activation functions for all the layers other than the last layer are ReLu activation. Final layer has a Linear Activation function. This network uses Mean Square Error(MSE) Loss function.

\subsection{Refiner Network}\label{Refiner-Network}
\begin{figure}
    \centering
    \includegraphics[width=5in]{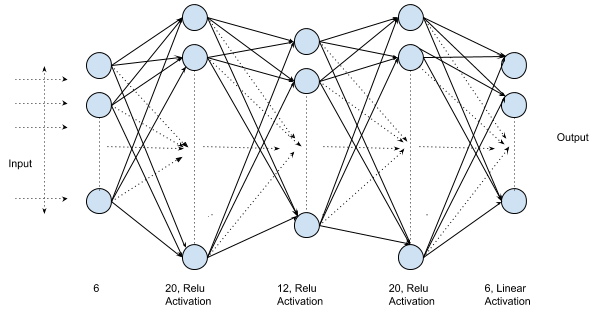}
    \caption{Refiner Network}
    \label{Refinement-Network}
\end{figure}

The architecture of Refiner Network is shown in Fig. \ref{Refinement-Network}. This network is a simple neural network with input corresponding to the output of CNN Network and labels as the pixel coordinates of the shoulder joint, elbow joint and wrist joint respectively. Activation function for the last layer is Linear and for rest is ReLu Activation. This network uses customized loss function that includes MSE and simple error in link length.

\subsection{Customized Loss Function}
For the error in coordinates of joint Mean Square Error(MSE) has been used and another error for link length has also been taken into consideration. Let $d$ denote error in link length and $p$ denote the MSE in position. For real joint coordinates $(x_s, y_s)$,  $(x_e, y_e)$,  $(x_w, y_w)$, and predicted coordinates $(x_{sp}, y_{sp})$,  $(x_{ep}, y_{ep})$,  $(x_{wp}, y_{wp})$ where subscripts $s$, $e$ and $p$ represents shoulder, elbow and wrist joints respectively and subscript $s$, $e$, $w$ followed by $p$ represents corresponding predicted coordinates respectively. The respective error are computed following the Eqns. (\ref{distd})-(\ref{loss}).

\begin{multline}\label{distd}
d = (\sqrt{(x_s - x_e)^2 - (y_s - y_e)^2} - \sqrt{(x_{sp} - x_{ep})^2 - (y_{sp} - y_{ep})^2}  )^2 + \\ (\sqrt{(x_e - x_w)^2 - (y_e - y_w)^2} - \sqrt{(x_{ep} - x_{wp})^2 - (y_{ep} - y_{wp})^2}  )^2
\end{multline}

\begin{equation}\label{distp}
  p = (x_s - x_{sp})^2 + (y_s - y_{sp})^2 + (x_e - x_{ep})^2 + (y_e - y_{ep})^2 + (x_w - x_{wp})^2 + (y_w - y_{wp})^2  
\end{equation}
\begin{equation}\label{loss}
   loss  = d + p  
\end{equation}

\subsection{Dataset}\label{dataset}
For the purpose of training the deep CNN, we created our own dataset of 300 images and manually labelled the pixel coordinates of the joints for each of the images. From the total dataset, 224 images were used for training and rest were used for testing.

\subsection{Training}
Our second approach requires training for which training and validation dataset has been used as specified under sub-heading \ref{dataset}. For the optimization purpose Adam optimizer has been used, with batch size of 14 images. The training has been done on the sample dataset for the case of one-link and two-link cases( forearm and arm) both on CNN network and Refinement network and the combined network. CNN network has been trained on 10 epoch and Combined Network on 100 epoch.

\section{Experiment}\label{exps}
The proposed method is tested with both real-time and recorded video sequences. But the processing of both type of the sequences are carried out in real-time since the 'mimic' of robotic arm is a real-time task. The experiments adressing four different challenges viz., i)presence of different level of background noise, ii) variation in distance between arm and camera, iii) variation in speed of hand movement, and iv) different number of links are performed to test the proposed methodology and setups are accordingly made. The experiments has been performed at frame rate of 30 fps and video resolution of 240x320 pixels.

\subsection{Experiment-1}
In the first experiment, the recorded video of hand motion is given as input to the network and the motion of the arm is mimicked by one-link manipulator in virtual environment of PyBullet. This experiment requires preparation of virtual environment and no physical setup is required. For the recorded video even noisy data was used to test the method. The recorded data used for testing is of \cite{Chalearn_dataset} (lossy compressed AVI format devel-1). Additional setups were not made for getting the recorded video. Videos used are available here as (
\href{https://drive.google.com/file/d/1NFrwCM8bgpx-QrJR9BCIeIOelafgKTom/view?usp=sharing}{Exp-1 Video-1}, \href{https://drive.google.com/file/d/1BoLfySqLC6fykoSk4Y0f12WjRcWp2kl0/view?usp=sharing}{Exp-1 Video-2})

\subsection{Experiment-2}
In the second experiment, the motion of actual one-link manipulator against stationary background is mimicked in simulation environment. This required preparation of physical setup of the manipulator and camera. The experimental setup consists of a camera(Model: HP HD 4310 H2W19AA) is mounted(fixed) at a height of 32 cm(can be varied) above the manipulator. The manipulator’s link length is 10.5 cm which is made up of paper to reduce the weight of the link mounted on the servo motor(specifications: Model: SM-S2309S, Size: $22.9\times12.3\times22.2$mm, Weight: 9.9g, Rotation angle $\equiv 120$ , Micro analog servo, 4 plastic gears + 1 metal gear). An Arduino UNO board has been used as a controller to provide signal to the servo motor for the motion (Fig.2(a)). The videos of experiment is available here as ( \href{https://drive.google.com/file/d/149BgtmAqipzvQsOHS-ClTN2R3aWwbNdc/view?usp=sharing}{Exp-2 Video-1}, \href{https://drive.google.com/file/d/1cdjHpQhrgU5Qbe78F8i3QuNJaZ9RO2F6/view?usp=sharing}{Exp-2 Video-2})

\begin{figure}[ht]
\includegraphics[width=2.5in]{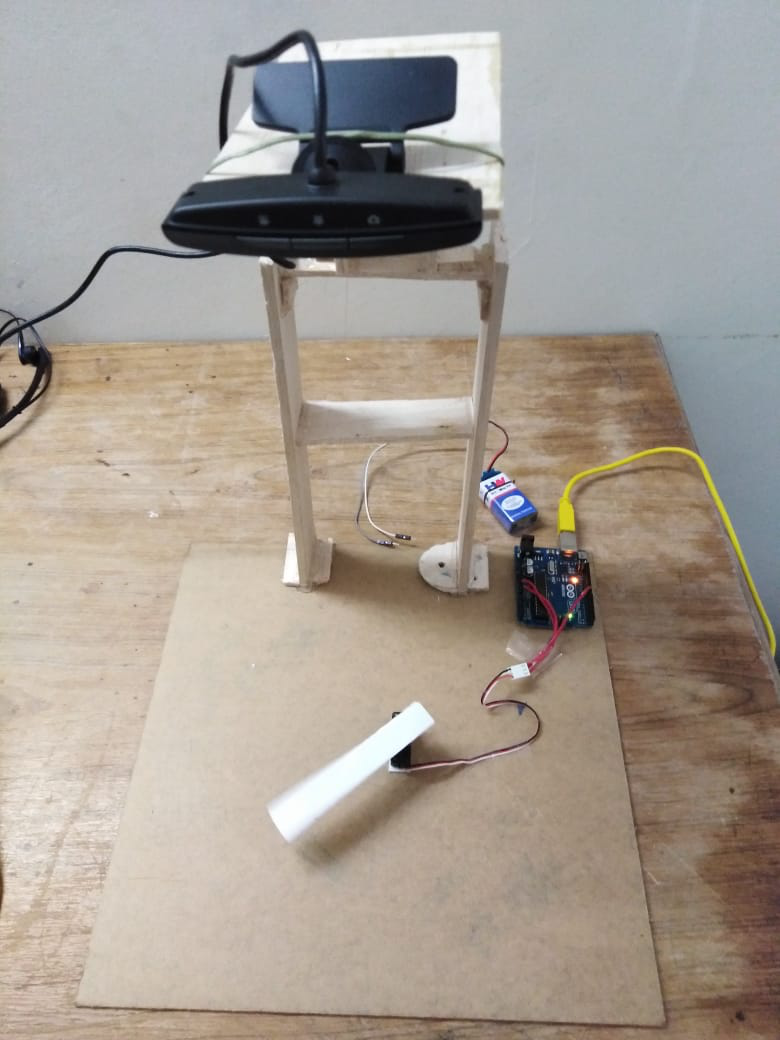}~\textbf{(a)}
\includegraphics[width=2.5in, height=2.0in]{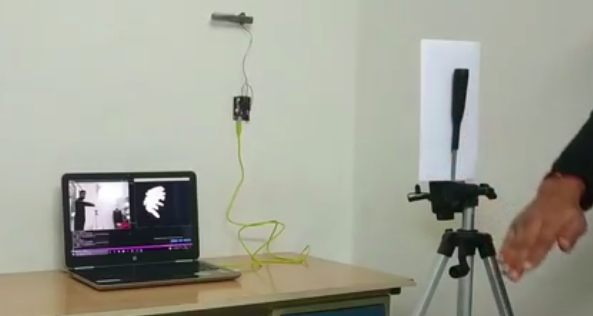}~\textbf{(b)}
\includegraphics[width=2.5in]{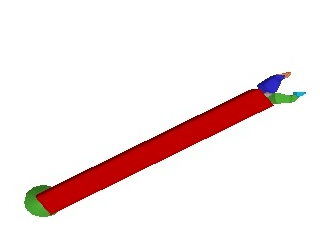}~\textbf{(c)}
  \caption{(a) Experimental setup for experiment-2, (b) Experimental setup for experiment-3, (c) Manipulator for experiment-4}
\end{figure}
\vspace{5cm}

\subsection{Experiment-3}
In the third experiment, the proposed method is used to mimic the forearm motion of a standing person by an actual one-link manipulator. The one-link manipulator used in Experiment-2 has been used here except the positioning has been changed as shown in Fig.2(b). Here, the distance between the forearm and the laptop(HP laptop AU030WM Pavilion) camera is varied. whereas the camera was mounted at fixed distance from the manipulator in earlier setup. Also, the background here is not stationary as noise is introduced while moving forearm other body parts too move slightly.

\subsection{Experiment-4}
In the fourth experiment, we extend our method to n-link planar manipulator(four-link, one-link is fixed). Here in case of human arm, we considered forearm, thumb and index finger having three joints overall. One additional layer to DON is added for the sake of experimentation here. It is shown in Fig. \ref{TR}. Please note no physical setup is prepared for this experiment and the testing is done in simulation. The manipulator used in Pybullet is shown in Fig.2(c)

\subsection{Parameter Tuning}
The actuator of our experimental setup enabled us to test the method using position control. However, velocity control and torque control techniques can also be used with the values obtained from the algorithm with the actuators that enables velocity control and torque control. For the uniformity, we use position control in simulator as well as on actual manipulator.

The joint angle computed in optical flow and the actual joint angle remains the same. Same is not the case with angular velocity, angular acceleration, torque and the link length. The term aspect ratio(ratio of value in optical flow and actual value) has been introduced for mapping optical flow value to actual value. These values shall be experimentally determined and is dependent on the experimental setup. 

Since the algorithm is completely unsupervised, the requirement of labelled data and domain adaptation is not required. The issue of cost is also dealt since the algorithm can easily run on low computation powered devices such as mobile handset, laptops, computers, etc. There is only lag time between the input of the image frame from video and the output signal to the manipulator which is the processing time.

\begin{figure}[ht]
\includegraphics[width=2.5in]{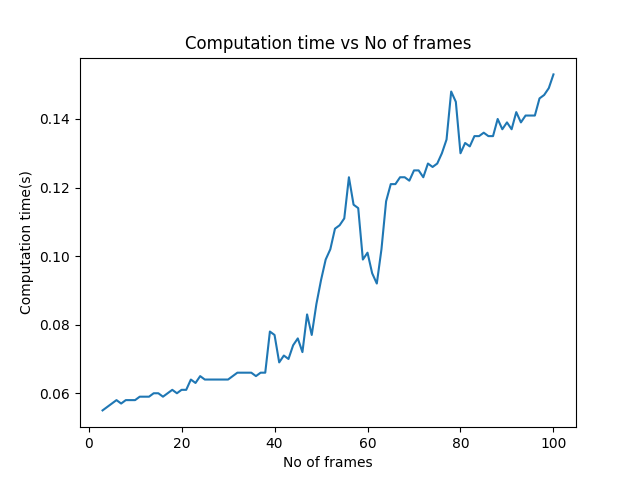}
\includegraphics[width=2.6in]{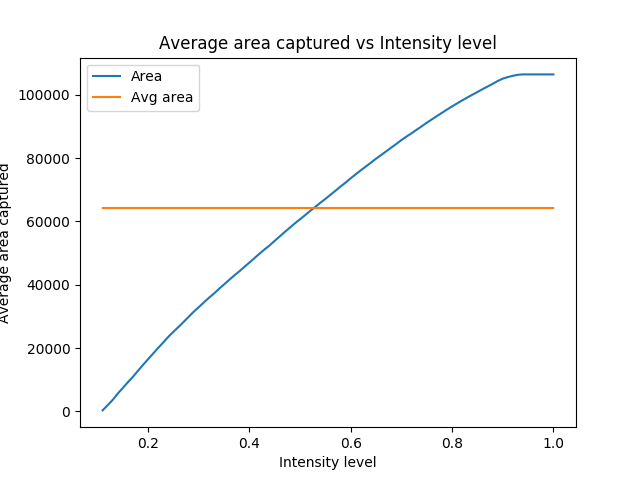}\\
\textbf{\noindent\hspace{2.5in} (a) \hspace{2.5in} (b) }
  \caption{(a) Computation time vs No of frames considered, (b) Average area captured vs intensity level}
\end{figure}

This processing time is a function of number of input frames (Fig.3(a)) in Input Layer, filters used, filter sizes, etc. The lag time can be decreased by using optimal number of input frames in Input Layer and appropriate filter with optimal filter size. For the case of experimentation the number of input frames is $10$ with Gaussian filter size $15$x$15$ which has been determined as optimal values through an iterative process.

Further the performance of Object Identification Layer is also dependent on intensity of light and background noise(movement of other objects). The algorithm works well for the intensity of light above 50\%(determined using experimental setup in Fig.2(a)) shown in Fig.3(b) and small background noise. The algorithm captures the major motion in the video. So, the performance of the algorithm is not affected until the major motion is of the arm.

\section{Results and Discussions}\label{Res}
\subsection{Results of DON}
The Experiment-1 performed on hand motion data set could hardly track the arm joint angle due to extremely random and fast hand movement and very large noise due to movement of other body part. This experiment was performed on both the RGB video and depth video. The results are available here RGB videos(\href{https://drive.google.com/file/d/11bgGX2z0_sMd-cZfif4E2Ep3vBdZzMu9/view?usp=sharing}{Video-1 at 10fps}, \href{https://drive.google.com/file/d/17tEPqNrmLmRaSaPcWeMEqGqHWeiwwNBa/view?usp=sharing}{Video-2 at 30fps}) and Depth videos (\href{https://drive.google.com/file/d/1XAAzrDJ32FX2pU5HceTbaZ96HERJ05CA/view?usp=sharing}{Video-1 at 10fps}, \href{https://drive.google.com/file/d/1Avm1DmR-S89xVtAxHryUrDIszM3kVsl_/view?usp=sharing}{Video-2 at 30fps}) \\

The Experiment-2 performed for mimicking actual one-link manipulator motion in simulation result is shown in Fig.4. In addition we used \cite{monodepth2} to generate depth images from RGB image and tested our algorithm. In both the cases our result obtained is the same.

\begin{figure}[ht]
\includegraphics[width=5in]{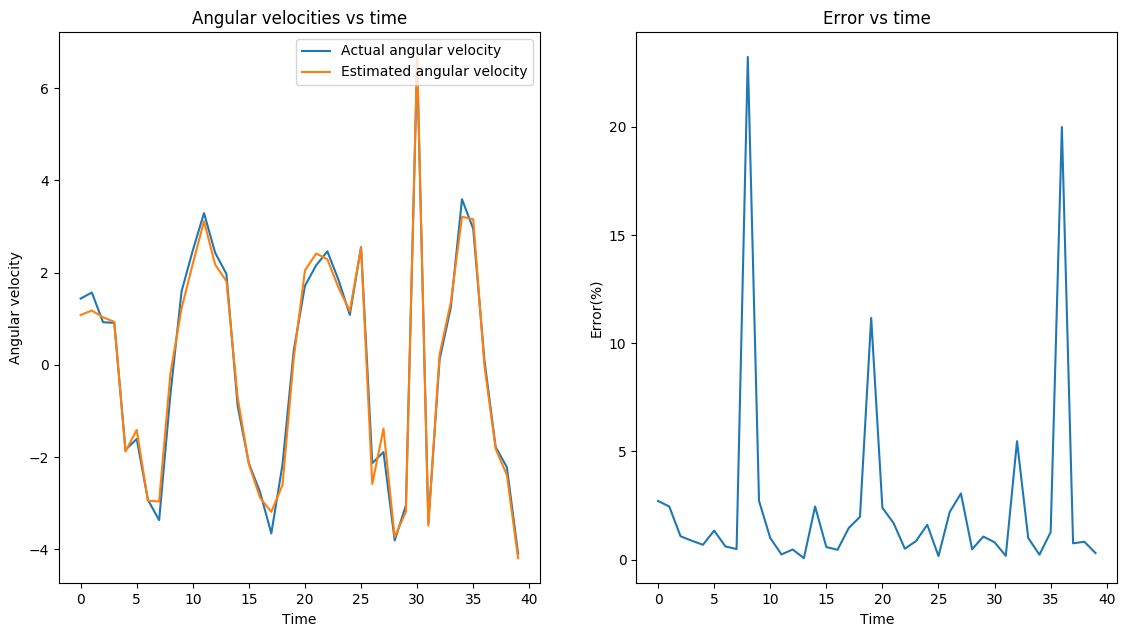}
\textbf{\noindent\hspace{2.5in} (a) \hspace{2.5in} (b) }
  \caption{Plot of (a)Angular velocity vs Time, (b) Error vs Time}
\end{figure}

\begin{figure}[ht]
\includegraphics[width=5in, height=2.0in]{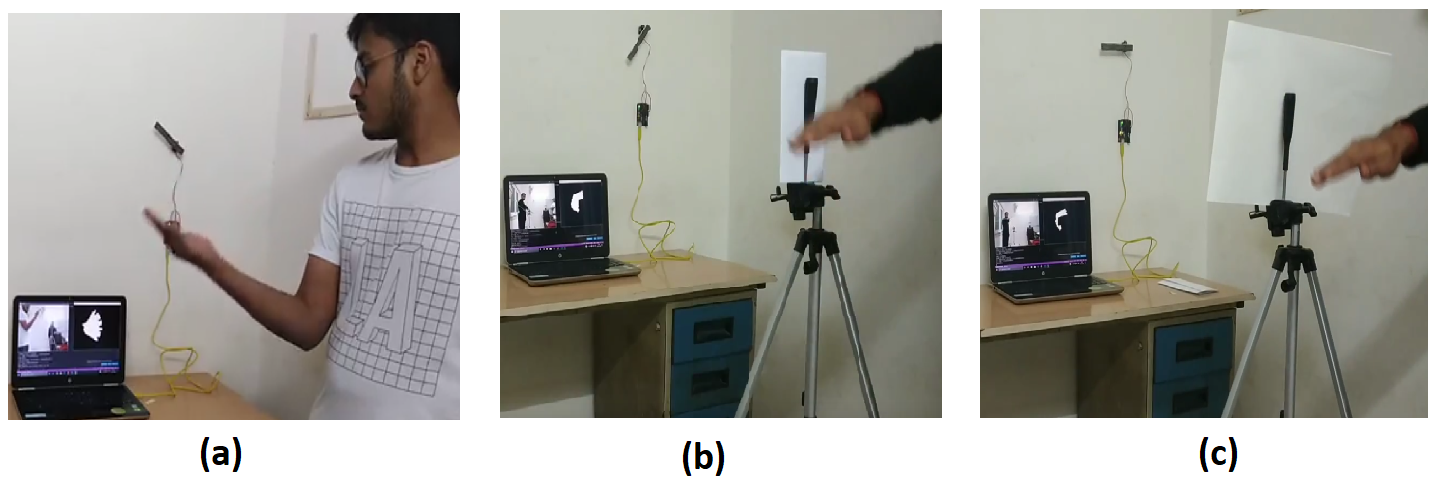}
  \caption{Visual \emph{i}-mimic in Real-time: (a) Without occlusion ($0.9<\eta<1$), (b) Low occlusion to background and human hand ($0.7<\eta<0.5$) and (c) High occlusion ($0.4<\eta<0.3$)}
  \label{TR}
\end{figure}

The Experiment-3 is performed with variation of trust factor ($\eta$ in Eqn. (\ref{TF})) that is with various degrees of occlusion. Example frames with arm-mimic for three different type of occlusions are shown in Fig.\ref{TR}, where there is no occlusion present there in Fig. \ref{TR}(a), low amount of occlusion is present there in Fig. \ref{TR}(b) and the amount of occlusion is quite high in Fig. \ref{TR}(c). It is also observed that the proposed method works well for moving hand. The videos corresponding to the experimental results are: \href{https://drive.google.com/file/d/1XZkAy8CRdnUxL03T6jsoyS19EZJ_umLj/view?usp=sharing}{no occlusion}, \href{https://drive.google.com/file/d/1fHxnL8go4Yo6iRDd618Jogb8KzEq9iDq/view?usp=sharing}{minor occlusion} and \href{https://drive.google.com/file/d/1fKvR7tOQGr4nfuCXO-ur0UaolWkpcDsC/view?usp=sharing}{major occlusion}.

\begin{figure}[ht]
\includegraphics[width=5in, height=2.0in]{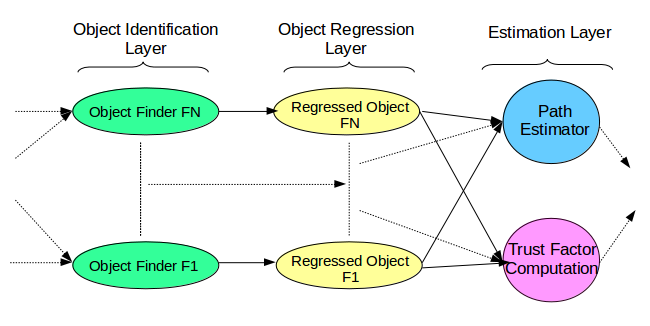}
  \caption{Adjusted portion of Architecture of Deep Flow Network}
  \label{TR}
\end{figure}

In Experiment-4, since there are three rotating link, three-joints angles are to be estimated. To accommodate multiple links, in the architecture of DON, an additional layer is added between Object Identification Layer and Estimation Layer to fit straight lines(Object Regression Layer) on objects as shown in Fig.6. Rest of the network remains the same. The joint angle between forearm and fixed link, thumb and forearm, index-finger and forearm are estimated accordingly. The results obtained after Object Regression Layer is shown in Fig.7.

The video expressing the obtained result is \href{https://drive.google.com/file/d/1I7184geEx9TTOOplQd9V-AOOJSHL7hIG/view?usp=sharing}{here}.

\begin{figure}[ht]
\includegraphics[width=5in, height=2.0in]{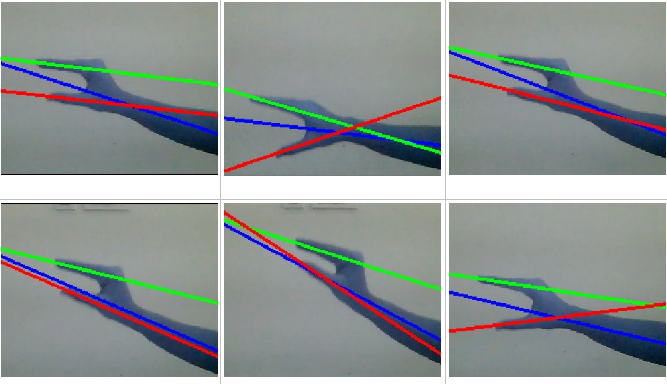}
  \caption{Snapshot image after Object Regression Layer}
  \label{TR}
\end{figure}

In addition to, the algorithm has been tested on videos of different resolutions at 30 fps. The average algorithm run time per loop execution for video input of different resolution before giving signal to manipulator for is presented in Table 1. This experiment has been performed on DELL Laptop with 8 GB RAM and Intel core i5, the algorithm run time will vary depending upon the computation power of hardware used for testing. 

\begin{table}[h!]
  \begin{center}
    \caption{Average loop run time for videos of different resolutions.}
    \label{tab:table1}
    \begin{tabular}{|c|c|} 
    \hline 
      \textbf{Resolution} & \textbf{Average loop run time(in
      milliseconds)}\\
    \hline 
      240x320 & 20\\
      480x640 & 45\\
      720x960 & 68\\
      960x1280 & 124 \\
    \hline
    \end{tabular}
  \end{center}
\end{table}

\subsection{Results of DNN}
\begin{figure}
    \centering
    \includegraphics[width=5in]{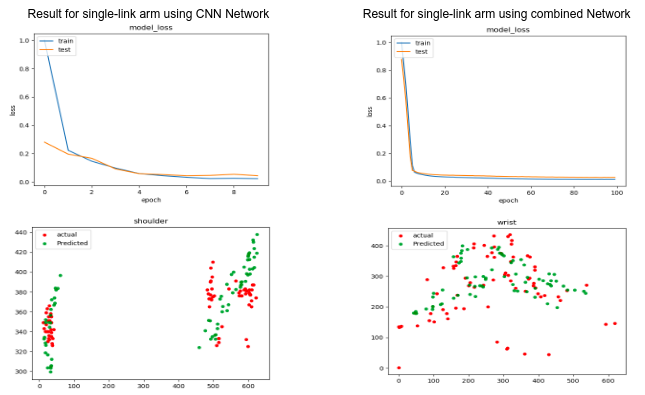}
    \caption{Result for single link arm}
    \label{res-single-link-arm}
\end{figure}

\begin{figure}
    \centering
    \includegraphics[width=5in]{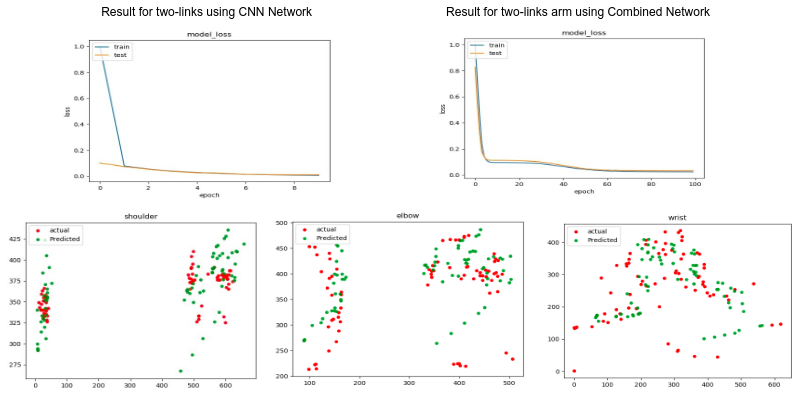}
    \caption{Results for two link arm}
    \label{res-two-link-arm}
\end{figure}

The CNN network and Refiner network has been tested on single link and two link cases. The loss occurred during the training  are shown in Fig. \ref{res-single-link-arm} and Fig. \ref{res-two-link-arm}. Likewise, the scattered plots of actual coordinate and predicted coordinates by combined Networks of one-link,  shoulders joint and wrist joint obtained are also shown in Fig. \ref{res-single-link-arm} and Fig. \ref{res-two-link-arm}.

\subsection{Discussions}
As we stated earlier, four different experiments under different circumstances, joints are performed here. In case of extremely random and very fast hand movement, the method is found to be ineffective as in Experiment-1 while in Experiment-2 where the background noise, arm speed are limited, the method performance i.e. mimicking is near to perfect. There is negligible lag time because the arm control command is given to simulator which runs nearly at 240Hz in Pybullet. In Experiment-3, the mimicking is performed with small lag time. This lag time(as seen in video) has been caused due to hardware limitation of the manipulator and setup. This can be reduced with good enough hardware, since the algorithm has been fine-tuned with optimal parameter values. The algorithm extended to n-link planar manipulator in Experiment-4 is able to estimate the three-joint angles between lines accurately as depicted in Fig.6. The algorithm can be extended to n-link planar manipulator just by introducing Object Regression Layer and the outcome would be as desired. Further, experimented with videos of different resolution shows the algorithm run time increases with increase in video resolution. 
In addition, our algorithm tracks/detects the objects on the basis of motion and not probabilistic color distribution or the object features. This enables our algorithm to run on both RGB and depth images independently and give the same result. \\

In case of our second method, for the one-link(forearm) case the Predicted Joint positions improved after adding a second network. The second network has acted as a smoothing/refining network which refines the outcome of the first network. For the two link cases, the performance of the combined network is better compared to just use of CNN Network.

\subsection{Comparative Study}

Please note that, first approach's application that we proposed here is new to literature. Therefore, no similar method is available to compare with. Therefore, we focus on comparing the proposed tracking algorithm. We could not conduct any direct comparative study for our tracking method too since no other tracking algorithm, formulated so far, gives torque and velocity as the output. For example, we carried out the same experimentation with two other robust and popular unsupervised tracking algorithms, namely, MoG2 and CAMSHIFT for the sake of comparison.
In this study we verified that the proposed algorithm tracks down the object within about 30 ms, whereas the time consumed by CAMSHIF and MoG2 are about 149 ms and 100ms respectively on HP laptop AU030WM Pavilion. Besides, these methods is not robust enough since they loose the object trajectory even to stationary background, therefore, its performance get reduced with reduction of trust factor.  Above all, none of the algorithm enables us to compute values of kinematic and dynamic parameters of motion like our algorithm for mimicking and hence failed to mimic. This is the main cause why we failed to carry out suitable comparative study for this application. 

For the second approach of deep learning, we compare our results to that of \cite{toshev2014deeppose} using Percentage correct parts(PCP) at link length threshold of 0.5(PCP 0.5), for upper arm and lower arm. PCP 0.5 was calculated on our model using our dataset on 20 images and also on 40 images. The comparison results are shown in Table \ref{tab:table2} and \ref{tab:table3} respectively. The data taken from \cite{toshev2014deeppose} is generalised on large dataset that has been used, for our case the method has not been generalised and results shown are obtained on our dataset.

\begin{table}[h!]
  \begin{center}
    \caption{Comparison PCP(0.5) on 20 images}
    \label{tab:table2}
    \begin{tabular}{|c|c|c|} 
    \hline 
      \textbf{Model} & \textbf{Upper Arm}  & \textbf{Lower Arm}\\
    \hline 
      Deep Pose 1st & 0.5 & 0.27\\
      Deep Pose 2nd & 0.56 & 0.35\\
      Deep Pose 3rd & 0.56 & 0.35\\
      CNN Network & 0.40 & 0.1 \\
      CNN + Refinement Network & 0.7 & 0.35 \\
    \hline
    \end{tabular}
  \end{center}
\end{table}

\begin{table}[h!]
  \begin{center}
    \caption{Comparison PCP(0.5) on 40 images}
    \label{tab:table3}
    \begin{tabular}{|c|c|c|} 
    \hline 
      \textbf{Model} & \textbf{Upper Arm}  & \textbf{Lower Arm}\\
    \hline 
      Deep Pose 1st & 0.5 & 0.27\\
      Deep Pose 2nd & 0.56 & 0.35\\
      Deep Pose 3rd & 0.56 & 0.35\\
      CNN Network & 0.25 & 0.125 \\
      CNN + Refinement Network & 0.566 & 0.275 \\
    \hline
    \end{tabular}
  \end{center}
\end{table}

From the above table, it is clear that our combined network comprising of CNN and Refiner Network performs better compared to just CNN network and Deep Pose Network for Upper Arm. However for lower the results are as good as that of Deep pose for the case of 20 images. On increasing the number of images to 40, performance decreases but still is better than that of Deep Pose  and CNN Network for Upper Arm. However, for lower arm combined is poor.

\section{Conclusions and Future Work}\label{con}

In the proposed work we aimed to develop a method with which robotic arms could be controlled only by showing video sequences in real-time. It is tested both with unsupervised DON network and deep CNN with manually labelled training samples. This approach is proven to be successful with adequate amount of demonstration shown here. The unsupervised DON based method is proved to be effective in achieving control over n-link manipulator in 2-D plane. Control over n-link manipulator plane can be achieved better with hybridization of CNN and refinement network as shown here in the study. The proposed technique performs well both with real-arm manipulator and synthetic arms. The mimic-based control therefore could be implemented to control robotic arm in different tasks. The approach is in its entry level as of now and more complex scenarios could be addressed in future with mimicking the motion of more than one joint on manipulator in spatial 3D environment.
\bibliographystyle{elsarticle-num}
\bibliography{ijcai20}
\end{document}